# Functional connectivity patterns of autism spectrum disorder identified by deep feature learning

Hongyoon Choi

*Abstract*— Autism spectrum disorder (ASD) is regarded as a brain disease with globally disrupted neuronal networks. Even though fMRI studies have revealed abnormal functional connectivity in ASD, they have not reached a consensus of the disrupted patterns. Here, a deep learning-based feature extraction method identifies multivariate and nonlinear functional connectivity patterns of ASD. Resting-state fMRI data of 972 subjects (465 ASD 507 normal controls) acquired from the Autism Brain Imaging Data Exchange were used. A functional connectivity matrix of each subject was generated using 90 predefined brain regions. As a data-driven feature extraction method without prior knowledge such as subjects' diagnosis, variational autoencoder (VAE) summarized the functional connectivity matrix into 2 features. Those feature values of ASD patients were statistically compared with those of controls. A feature was significantly different between ASD and normal controls ($p < 1 \times 10^{-6}$). The extracted features were visualized by VAE-based generator which can produce virtual functional connectivity matrices. The ASD-related feature was associated with frontoparietal connections, interconnections of the dorsal medial frontal cortex and corticostriatal connections. It also showed a trend of negative correlation with full-scale IQ. A data-driven feature extraction based on deep learning could identify complex patterns of functional connectivity of ASD. This approach will help discover complex patterns of abnormalities in brain connectivity in various brain disorders.

## I. INTRODUCTION

Autism spectrum disorder (ASD) is a group of developmental disabilities characterized by impaired social and communication skills combined with repetitive behaviors and fixated interests [1]. Current diagnosis of ASD is primarily based on assessing the behavioral characteristics and intellectual properties as brain abnormalities and neural mechanisms of ASD remain largely unknown. A recent hypothesis of the neural basis of ASD is that abnormalities are associated with disrupted neuronal connections. It has been recognized that ASD has disrupted functional connectivity between the multiple brain regions, eventually affecting global brain networks [2, 3]. In particular, abnormal interaction across brain networks in ASD disrupts local neuronal circuitry, which causes local over-connections and reduce long-range connections [4, 5].

To evaluate functional brain network, functional magnetic resonance imaging (fMRI) has been widely used. Functionally connected brain regions can be identified by synchronized neural activity which can be detected as fluctuations in Blood Oxygen Level Dependent (BOLD) signal. Using resting-state fMRI, several studies have reported that ASD has abnormal connectivity in long-range connections [6-9]. However, they have not reached a consensus of disrupted functional networks in accordance with which connections cause the ASD-related symptoms and whether they are increased or decreased [10, 11]. A possible cause of the difficulties in identifying abnormal connections of ASD is that previous studies have generally compared specific networks or connectivity between individual brain regions. Group differences of connections between specific brain regions hardly support generalized and complex abnormal patterns. As ASD affects multiple functional connections, neuronal activity identified by fMRI can show complicated patterns of abnormal connections across the entire brain rather than abnormality in some specific regions. Therefore, to understand this functional connectivity pattern of ASD in large-scale, the evaluation of multivariate features is needed.

We aimed to find multivariate and nonlinear patterns of resting-state functional connectivity in ASD using unsupervised feature extraction and dimension reduction. Unsupervised learning can learn features of high-dimensional data without prior information [12, 13]. As a type of deep learning, autoencoder with variational Bayes is an unsupervised learning system which can extract features of data without label and, even more, generate virtual data from the features [14]. We applied this variational autoencoder (VAE) to summarize multivariate and high-dimensional data into 2-dimensional features. Consequently, we investigated which functional connectivity pattern is associated with ASD.

.

This work was supported by funding for ABIDE listed at http://fcon_1000.projects.nitrc.org/indi/abide/

Hongyoon Choi, MD.,Ph.D. is with Department of Nuclear Medicine, Cheonan Public Health Center, 234-1 Buldang-Dong, Seobuk-Gu, Cheonan, Republic of Korea. (E-mail: chy1000@gmail.com)

## II. MATERIALS AND METHODS

### A. Materials

We used data acquired from the Autism Brain Imaging Data Exchange (ABIDE) repository of resting-state fMRI scans [15]. fMRI scans of this repository were obtained from multiple international sites (http://fcon_1000.projects.nitrc.org/indi/abide/ for more information). It also includes subjects' diagnosis as well as various measurements related to ASD. All data were anonymized in accordance with Health Insurance Portability and Accountability Act (HIPAA) guidelines. All data were based on ABIDE studies approved by local Institutional Review Board (IRB). We used publically available preprocessed version of ABIDE dataset for easy replication of our approach [16]. 972 subjects (465 ASD patients and 507 normal controls, NCs) with complete fMRI and diagnostic data were selected. Demographic data is shown in Table 1.

### B. Data processing

We used preprocessed data with Data Processing Assistant for Resting-State fMRI (DPARSF) pipelines [16]. Resting-state fMRI data preprocessing procedures included slice-time correction, time-series image realignment using rigid-body transformation, and head-motion correction using 24-parameter model [17]. The first four volumes of time-series data were discarded before the preprocessing steps. Nuisance regression was performed using the signals from white matter, cerebrospinal fluid and global signal. Linear and quadratic trends of BOLD signal which represented low-frequency drifts were also regressed out. After the regression, temporal filtering (0.01 - 0.1 Hz) was performed.

Regional BOLD signal fluctuations were measured by using Automated Anatomical Labeling (AAL) template [18]. The template included 45 brain regions for each cerebral hemisphere (Table 2 with abbreviation of each brain region). To evaluate functional connectivity between brain regions, absolute values of Pearson correlation coefficients were used for edge weights of subjects' brain graph. Each brain region was regarded as a node the graph. The edge weight of each subject was used for the further analysis.

TABLE I
DEMOGRAPHIC DATA OF SUBJECTS

|  | Autism spectrum disorder | Normal control |
|---|---|---|
| Number of subjects | 465 | 507 |
| Age | 16.5±7.9 (7.0-58.0) | 16.5±7.2 (6.5-56.2) |
| Sex (M:F) | 409/56 | 414/93 |
| full-scale IQ | 105.1±16.9 | 110.7±12.5 |
| Diagnosis* | Autism: 295 Asperger: 85 PDD-NOS: 3 Asperger of PDD-NOS: 1 | - |

* 81 subjects without subtype information

TABLE II
ABBREVIATION FOR BRAIN REGIONS

| Abbreviation | Full name of brain regions |
|---|---|
| PreCG | Precentral gyrus |
| SFGdor | Superior frontal gyrus, dorsolateral |
| ORBsup | Superior frontal gyrus, orbital part |
| MFG | Middle frontal gyrus |
| ORBmid | Middle frontal gyrus orbital part |
| IFGoperc | Inferior frontal gyrus, opercular part |
| IFGtriang | Inferior frontal gyrus, triangular part |
| ORBinf | Inferior frontal gyrus, orbital part |
| ROL | Rolandic operculum |
| SMA | Supplementary motor area |
| OLF | Olfactory cortex |
| SFGmed | Superior frontal gyrus, medial |
| ORBsupmed | Superior frontal gyrus, medial orbital |
| REC | Gyrus rectus |
| INS | Insula |
| ACG | Anterior cingulate and paracingulate gyri |
| DCG | Median cingulate and paracingulate gyri |
| PCG | Posterior cingulate gyrus |
| HIP | Hippocampus |
| PHG | Parahippocampal gyrus |
| AMYG | Amygdala |
| CAL | Calcarine fissure and surrounding cortex |
| CUN | Cuneus |
| LING | Lingual gyrus |
| SOG | Superior occipital gyrus |
| MOG | Middle occipital gyrus |
| IOG | Inferior occipital gyrus |
| FFG | Fusiform gyrus |
| PoCG | Postcentral gyrus |
| SPG | Superior parietal gyrus |
| IPL | Inferior parietal, but supramarginal and angular gyri |
| SMG | Supramarginal gyrus |
| ANG | Angular gyrus |
| PCUN | Precuneus |
| PCL | Paracentral lobule |
| CAU | Caudate nucleus |
| PUT | Lenticular nucleus, putamen |
| PAL | Lenticular nucleus, pallidum |
| THA | Thalamus |
| HES | Heschl gyrus |
| STG | Superior temporal gyrus |
| TPOsup | Temporal pole: superior temporal gyrus |
| MTG | Middle temporal gyrus |
| TPOmid | Temporal pole: middle temporal gyrus |
| ITG | Inferior temporal gyrus |

### C. Variational autoencoder model for connectivity data

In this study, we used VAE model to identify multivariate patterns of functional connectivity data. An overview of VAE-based ASD-related feature extraction is presented in Fig. 1. VAE is an unsupervised learning method that can compress the data to low-dimensional features. Without prior knowledge including subjects' diagnosis, functional connectivity matrices were used to train VAE model.

An autoencoder is an unsupervised learning method to learn a feature for a set of data and reduce data dimension. The basic form of autoencoder consists of encoder and decoder. Encoder reduces the dimension of data $X$ by compressing them into latent features $Z$. It consists of multiple layers of the neural network. Decoder reconstructs data $X$ from latent features. Connecting encoder and decoder, the input and output layer have same number of nodes.

VAE is additionally make assumptions concerning the distribution of latent variables. It can learn a probability distribution of data $x \in X$ in latent feature space. Thus, the encoded latent feature of VAE can provide a continuous variable representing a multivariate pattern of the original data $X$. More specifically, it is assumed that the data were generated from some conditional distribution and an unobserved continuous random variable $z$ in latent space $p_\theta(x|z)$. The encoder is an approximation $q_\phi(z|x)$ to the intractable true posterior $p_\theta(z|x)$. The objective function of variation autoencoder is

$$L(\phi, \theta) = -E_{z \sim q_\phi(z|x)}(\log p_\theta(x|z)) + KL(q_\phi(z|x) \| p_\theta(z)) \quad (1)$$

where $\theta, \phi$ represents generator and encoder parameters. The first term represents reconstruction loss, expected log-likelihood for the data. KL is Kullback-Leibler divergence between the learnt latent distribution the prior distribution $p_\theta(z)$, acting as a regularization term [14]. To train VAE, data were encoded into parameters in a latent space, and decoder network reconstructs data from the latent features assuming they have normal distribution around encoded feature $z$.

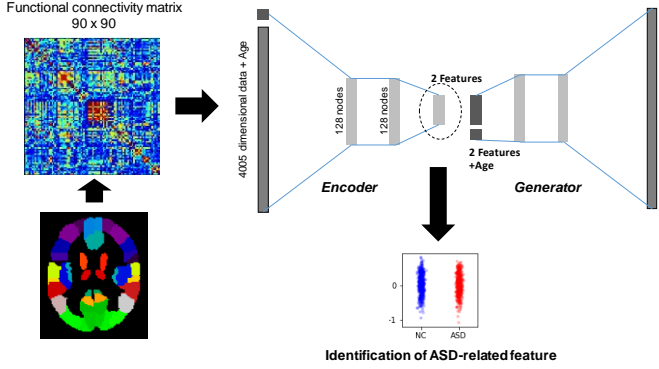

Fig. 1. Workflow for identification of ASD-related feature using variational autoencoder. Regional brain activity was calculated using predefined volume-of-interests, automated anatomical labeling (AAL). Functional connectivity matrix for each subject was generated based on Pearson correlation coefficients. The functional connectivity data of ASD patients and normal controls (NC) were entered into variational autoencoder (VAE) to extract data-driven features without prior information. 2 features were extracted and those feature values of ASD and NC were statistically compared.

Input data of VAE were connectivity matrices of subjects. Because of the symmetry of 90 × 90 edge weight matrix, upper triangular values of the matrix were entered into the VAE model. In addition, we added a dimension for age information to extract features independent from age. Input nodes with 4006 dimensions (4005 + age information) were connected to two hidden layers with 128 nodes. They were followed by latent feature layers with 2 dimensions. For decoding, generator input was randomly resampled by the encoded latent features $z$ assuming normal distribution: $z_{resampled} = z_{encoder} + z_{sd} \times \varepsilon$, where $\varepsilon$ represents a random variable [14]. The generator input was also merged with age information and it was connected to two hidden layers with 128 nodes. It was eventually connected to output layer that has 4005 dimensions.

Among the 972 functional connectivity data, 874 were used for the VAE training and 98 were used for the validation to monitor loss and determine termination of iterative training epochs. The training/validation set was randomly sampled keeping the same proportion of ASD though training process did not use diagnosis label. Our framework was trained by gradient descent algorithm (Adadelta)[19] and took 50 epochs for the training. The VAE was implemented using a deep learning library, Keras (ver. 1.2.2) [20]. After the training, encoder extracted 2 features of the functional connectivity data of 972 subjects.

### D. Identification of ASD-related feature

2 encoded features of ASD and NCs were compared using independent t-test. To define ASD-related feature, the feature significantly different between ASD and NCs were selected using p < 0.05. To investigate whether the extracted features were associated with clinical variables, the ASD-related feature also correlated with full-scale IQ test results using Pearson correlation.

As encoded features have multivariate and nonlinear properties, visualization of patterns in the original functional connectivity data is difficult. Instead, ASD-related features were indirectly visualized using generative decoder. We focused on how changes in ASD-related features affect functional connectivity data. As the generator decoded latent features to the original data space, it could generate virtual functional connectivity data according to the any values of latent features. Firstly, a representative functional connectivity data of all subjects, mean values of 2 features were inputted to the generative decoder: $p_\theta(X_{mean}|Z_{mean})$. To visualize an ASD-related feature ($z_i \in Z$), virtual connectivity data were generated according to the changed feature value from -1-standard deviation (SD) to 1-SD. We visualized changes in virtual functional connectivity data according to the changes in a given feature.

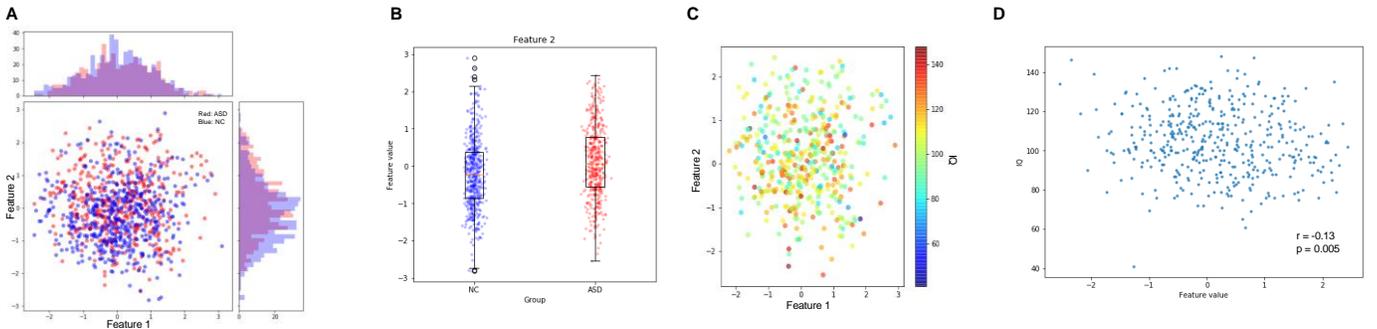

Fig. 2. Two features encoded by VAE. Encoder extracted 2 features from functional connectivity data of each subject. (A) A scatter plot with the two features was drawn. Red and blue dots represented ASD and NCs, respectively. (B) The ASD-related feature value of ASD was significantly lower than that of NC (0.12±0.97 vs. -0.20±0.94, p < 1×10$^{-6}$). (C) A scatter plot was drawn for full-scale IQ test scores. It exhibited a trend that data with lower IQ scores were located on upper portion. (D) The ASD-related feature was significantly correlated with IQ score (r = -0.13, p = 0.005).

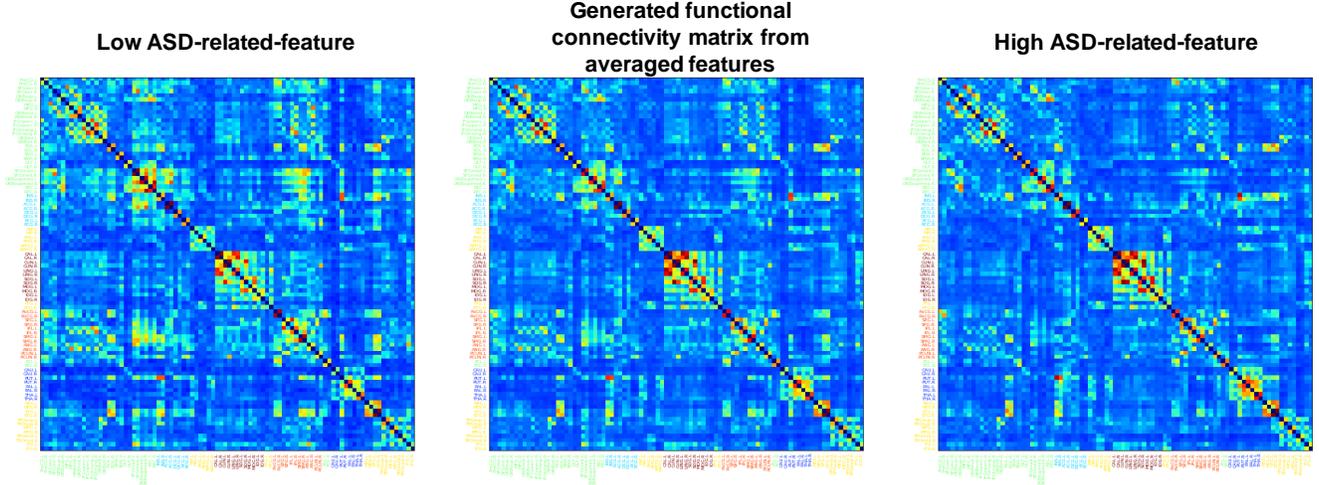

Fig. 3. Generated functional connectivity matrices. To understand the ASD-related feature, virtually generated functional connectivity matrices were drawn. At first, the average value of features across all subjects was entered into the generator. Additional functional connectivity matrices were generated by changing the value of ASD-related feature.

$Feature\ i$:
$$\Delta p(X|Z) \sim p(X|z_i = \mu_i \pm \sigma_i) - p(X|z_i = \mu_i). \quad (2)$$

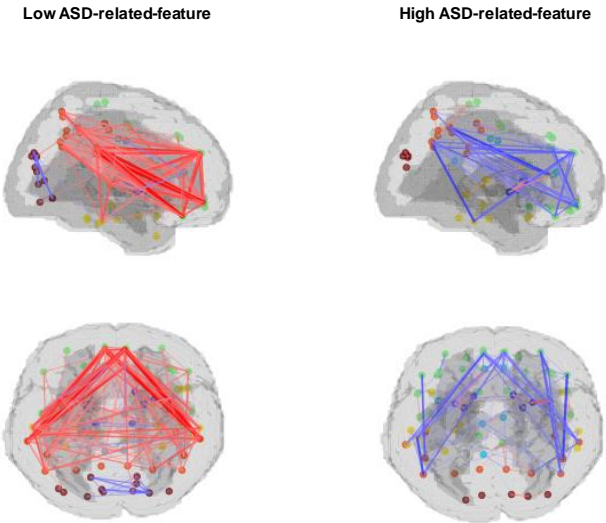

Fig. 4. Visualization of ASD-related feature. The generated functional connectivity matrices by low (mean – SD) and high (mean + SD) ASD-related feature value were subtracted by the matrix generated by the average value. Red and blue lines represented increased and decreased functional connectivity according to the subtraction. ASD-related feature was associated with regional connectivity in dorsal medial frontal cortices (superior frontal gyrus and gyrus rectus) and frontoparietal connections.

We also generated the manifold of functional connectivity patterns. Latent features were changed from -2 to 2 for both axes and connectivity features were drawn by subtracting the connectivity matrix generated by feature values = (0, 0).
In addition, functional connectivity strength (FCS) of 90 brain regions was evaluated. FCS was defined by sum of edge weight of each node. The changes in FCS according to the changes in ASD-related features were also evaluated.

## III. RESULTS

2 functional connectivity features for 972 subjects were extracted using VAE. According to the two features, functional connectivity data were plotted (Fig. 2A). A feature was significantly associated with ASD. The ASD-related feature was significantly increased in ASD compared with NCs (ASD-related-feature 1: 0.12±0.97 vs. -0.20±0.94 for ASD and NC, p < 1×10$^{-6}$) (Fig. 2B). ASD-related feature was significantly negatively correlated with full-scale IQ scores (r = -0.13, p = 0.005) (Fig. 2C, D).

ASD-related-feature was visualized by VAE-based generator for functional connectivity matrix. We generated virtual functional connectivity matrices according to changing ASD-related-features. They were generated using a low ASD-related feature value (mean – SD) and a high ASD-related feature value (mean + SD), respectively (Fig. 3). To find major changes in the functional connectivity matrices according to the ASD-related features, the difference of edges was redrawn. High ASD-related feature represented decreased regional connectivity in medial frontal cortices (superior frontal gyrus and gyrus rectus) and decreased interconnections between medial frontal and parietal cortices increased connectivity between subcortical structures and parietal lobes. Note that ASD-related feature was higher in ASD than NCs. It also represented increased connectivity of subcortical structures (Fig. 4, Fig. 5). Low ASD-related feature represents increased FCS in medial frontal, inferior frontal, posterior cingulate, precuneus, supramarginal and angular gyrus. High ASD-related feature was associated with increased FCS in the right putamen and pallidum and decreased FCS in the medial frontal and angular gyri (Fig. 6). The learned manifold exhibits globally interhemispheric interconnections were associated with the first feature (vertical axis) and frontoparietal connections were associated with the second feature (horizontal axis), which was ASD-related feature (Fig. 7).

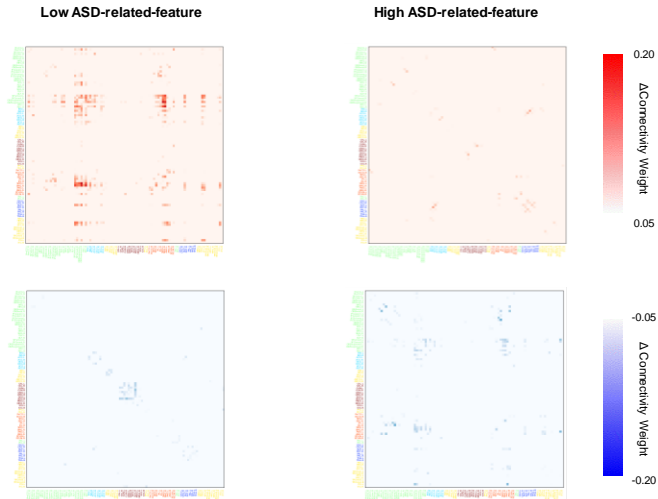

Fig. 5. Subtracted functional connectivity matrices generated by different values of ASD-related feature. Subtracted functional connectivity matrices were drawn. Virtual functional connectivity matrices were generated by low (mean – SD) and high (mean + SD) ASD-related feature. They were subtracted by the matrix generated by the average feature value to visualize the functional connectivity pattern represented by ASD-related feature.

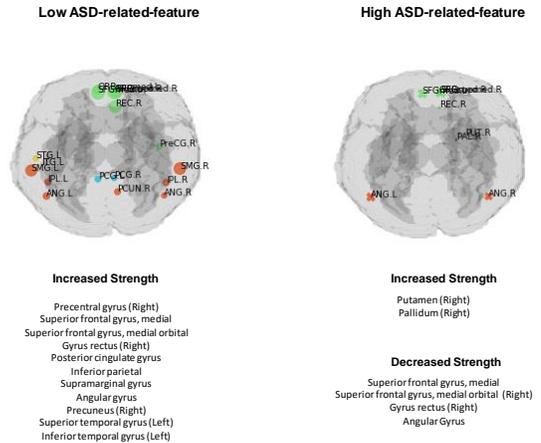

Fig. 6. ASD-related feature and functional connectivity strength. Changes in functional connectivity strength (FCS) according to the ASD-related feature were presented. Circle represented increased FCS and 'X' represented decreased FCS. Brain regions with ΔFCS > 1.5 were annotated.

## IV. DISCUSSION

In this study, we identified multivariate and nonlinear patterns of functional connectivity in ASD using features extracted by VAE. Abnormally synchronized neuronal activity in ASD is involved in multiple regions and could affect overall brain networks [2, 3]. Thus, group comparison focusing on individual connections of specific brain regions hardly identifies abnormal patterns of globally disrupted brain connectivity. It has caused mixed results of abnormal functional connectivity of ASD in previous studies [10, 11]. VAE could identify complicated patterns of functional connectivity concentrating on global patterns rather than individual connections of some specific regions. In addition, as an unsupervised learning, it could extract the features from data without any prior information.

We found that a feature was significantly associated with ASD. It was encoded by a deep neural network, which effectively compressed data to low dimensional space and extracted patterns of the data. However, intuitive visualization of the encoded feature is difficult [21]. To understand which patterns were associated with the feature, it could be indirectly identified by generating virtual connectivity matrices according to changing values of each feature. A recent study also used VAE to extract features from structural MRI data and visualized those features by changing feature values [22]. This approach could be used to represent complicated nonlinear patterns of brain imaging data identified by deep neural networks. Such a VAE-based feature extraction could apply to identification of various disease-related patterns in brain connectivity data.

The feature significantly associated with patients with ASD could reflect globally disrupted functional connectivity patterns. The ASD-related feature represented decreased interconnections of medial frontal cortex and frontoparietal cortex. It was also associated with increased subcortical connections. Various studies have focused on the characterized neuronal synchronization patterns in ASD summarized by decreased long-range connections with local overconnectivity [4, 5]. In particular, interconnections of frontoparietal cortex have been regarded as a representative regions of long-range underconnectivity [23-26]. High ASD-related feature was related to low FCS in dorsal medial frontal cortices as well as interconnections of the frontal cortex. It suggested ASD was associated with relatively low FCS in medial frontal cortices. Several neuroimaging studies have identified that the social disturbance of autism is closely associated with the dorsal medial frontal cortex [27-30]. In addition, as abnormal subcortical circuits in ASD have been identified several times [8, 31], the high ASD-related feature associated with increased FCS in subcortical structures reflected the previously reported aberrant subcortical connections. Accordingly, ASD-related feature included various functional connectivity patterns previously identified by studies in terms of crucial regional changes or specific interconnections.

In this study, the fMRI scans obtained from various centers were included, which could affect the feature extraction. Furthermore, various clinical factors could limit the robust feature extraction. Nonetheless, our model used age information as inputs so that the features were independent from the subjects' age which might affect functional connectivity patterns. A well-controlled prospective design to evaluate the result of ASD-related pattern will provide more precise patterns of ASD. Even though ASD patients showed significantly different ASD-related features, the values were considerably overlapped in both groups. The diagnostic accuracy for differentiating ASD from NC measured by area of receiver-operating-characteristics curve was 0.60 (Fig. 8). Therefore, in order to use these VAE-based features for diagnosis, additional modification and optimization will be needed. As the feature was obtained by the unsupervised learning at present, higher diagnostic accuracy could be achieved with a large cohort and further supervised training. Recently, several classifiers have been developed to differentiate ASD from NC using fMRI data [32, 33]. VAE-based features could apply to various types of classifiers for developing diagnostic biomarkers as a future study.

The functional connectivity patterns were obtained by neural activity of brain regions defined by AAL. As a deep learning-based approach can deal with high dimensional data to extract features, brain activity measurements with smaller predefined volume-of-interests or voxelwise analyses could improve the results. As a proof-of-concept study of VAE-based feature extraction in functional brain network, further works with feature extraction from higher dimensional data to find global connectivity patterns will help elucidate the pathophysiology of ASD.

V. CONCLUSIONS

As brain networks are globally disrupted in ASD, functional connectivity patterns of the patients are complex. VAE, a type of unsupervised learning for feature extraction, could identify multivariate patterns of functional connectivity related to ASD. This extracted pattern included brain regions repeatedly reported in previous studies with regard to social behavior. Furthermore, the features also corresponded to the hypothesis of brain network patterns of local overconnectivity and long-range underconnectivity. As several brain disorders affect multiple brain regions and global brain networks, a comprehensive pattern analyses will be needed to understand pathophysiology as well as to develop novel diagnostic biomarkers. This unsupervised learning-based feature extraction can be a useful method which helps identify complex patterns of brain networks in several neuropsychiatric disorders.

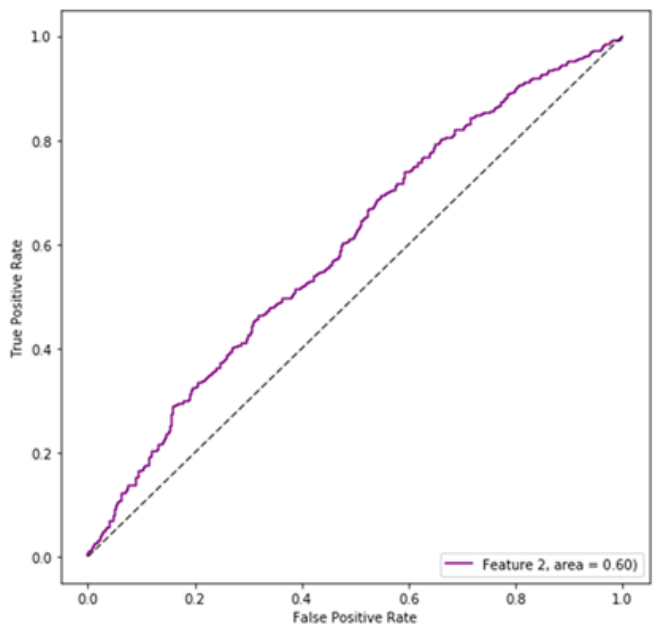

Fig. 8. Receiver-operating-characteristics curve using ASD-related feature. To evaluate diagnostic accuracy of ASD-related features, receiver-operating-characteristics curves were drawn for the ASD-related feature value. Area-under-curve was 0.60.

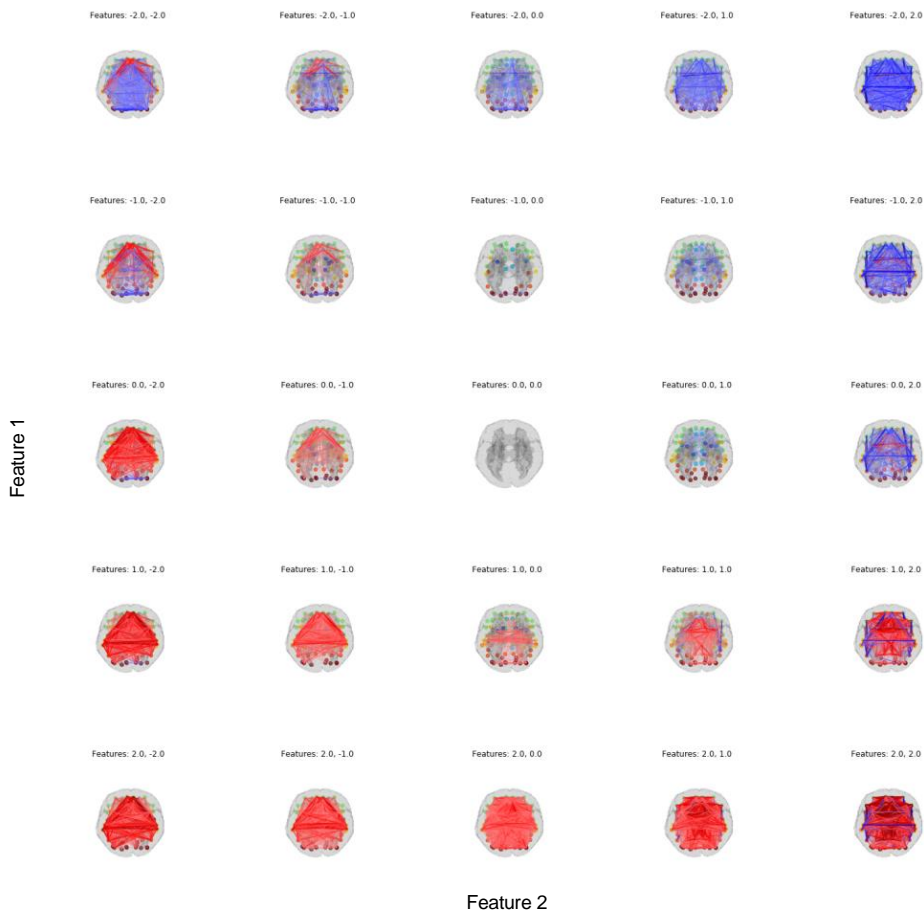

Fig. 7. Manifold of two features. Functional connectivity matrices were generated according to two feature values. They were subtracted by the functional connectivity matrix of feature values 0.

ACKNOWLEDGMENT

This work was supported by funding for ABIDE listed at http://fcon_1000.projects.nitrc.org/indi/abide/